\documentclass[12pt,a4paper,twoside]{article}
\makeatletter\if@twocolumn\PassOptionsToPackage{switch}{lineno}\else\fi\makeatother

\usepackage{amsfonts,amssymb,amsbsy,latexsym,amsmath,tabulary,graphicx,times,xcolor}
\usepackage[T1]{fontenc}
\usepackage{url,multirow,morefloats,floatflt,cancel,tfrupee,multicol}
\makeatletter
\usepackage[ruled,vlined]{algorithm2e}

\AtBeginDocument{\@ifpackageloaded{textcomp}{}{\usepackage{textcomp}}}
\makeatother
\usepackage{colortbl}
\usepackage{pifont}
\usepackage{floatrow}
\usepackage[nointegrals]{wasysym}
\usepackage[superscript,biblabel]{cite}
\usepackage{comment}
\usepackage{gensymb}
\usepackage[normalem]{ulem}
\makeatletter

\@ifundefined{etal}{\def\etal{\textit{et~al}}}{}

\AtBeginDocument{
\expandafter\ifx\csname eqalign\endcsname\relax
\def\eqalign#1{\null\vcenter{\def\\{\cr}\openup\jot\m@th
  \ialign{\strut$\displaystyle{##}$\hfil&$\displaystyle{{}##}$\hfil
      \crcr#1\crcr}}\,}
\fi
}

\@ifundefined{tsGraphicsScaleX}{\gdef\tsGraphicsScaleX{1}}{}
\@ifundefined{tsGraphicsScaleY}{\gdef\tsGraphicsScaleY{.9}}{}
\def\checkGraphicsWidth{\ifdim\Gin@nat@width>\linewidth
	\tsGraphicsScaleX\linewidth\else\Gin@nat@width\fi}

\def\checkGraphicsHeight{\ifdim\Gin@nat@height>.9\textheight
	\tsGraphicsScaleY\textheight\else\Gin@nat@height\fi}

\def\fixFloatSize#1{}
\let\ts@includegraphics\includegraphics

\def\inlinegraphic[#1]#2{{\edef\@tempa{#1}\edef\baseline@shift{\ifx\@tempa\@empty0\else#1\fi}\edef\tempZ{\the\numexpr(\numexpr(\baseline@shift*\f@size/100))}\protect\raisebox{\tempZ pt}{\ts@includegraphics{#2}}}}

\AtBeginDocument{\def\includegraphics{\@ifnextchar[{\ts@includegraphics}{\ts@includegraphics[width=\checkGraphicsWidth,height=\checkGraphicsHeight,keepaspectratio]}}}

\DeclareMathAlphabet{\mathpzc}{OT1}{pzc}{m}{it}

\def\URL#1#2{\@ifundefined{href}{#2}{\href{#1}{#2}}}

\edef\fntEncoding{\f@encoding}

\makeatother

\newif\ifmultipleabstract\multipleabstractfalse%
\usepackage[hmargin=1.8cm,vmargin=2.2cm]{geometry}
\makeatletter
\def\author#1{\gdef\@author{\hskip-\dimexpr(\tabcolsep)\hskip1pt\parbox{\dimexpr\textwidth-1pt}{\centering #1}}}

\let\@articletype\@empty \def\articletype#1{\gdef\@articletype{{\fontsize{14}{16}\selectfont #1}}}

\usepackage{fancyhdr}

\fancypagestyle{headings}{\fancyhf{}\fancyhead[RE]{\MakeTextUppercase{\textbf{\RunningAuthor}}}\fancyhead[LE]{\textbf{\thepage}}\fancyhead[LO]{\MakeTextUppercase{\textbf{\RunningHead}}}\fancyhead[RO]{\textbf{\thepage}}}
\fancypagestyle{plain}{\fancyhf{}\fancyfoot[C]{\textbf{\thepage}}}

\AtBeginDocument{\usepackage{lastpage}}
\def\title#1{%
  \gdef\@title{%
    \ifx\@articletype\@empty\else\@articletype~\\\fi%
     #1}%
}

\usepackage{abstract}
\def\abstractname{\textbf{Abstract}}
\renewenvironment{onecolabstract}
{\vspace*{-.4pc}\trivlist\item[]\leftskip1pt\noindent\selectfont\hfill\abstractname\hfill\mbox{\null}\par\ignorespaces}{\endtrivlist}

\def\NormalBaseline{\def\baselinestretch{1.1}}

\usepackage{textcase}
\usepackage[explicit]{titlesec}
\titleformat{\section}[block]{\NormalBaseline\boldmath\bfseries}
{\thesection.}
{6pt}
{#1}
[]
\titleformat{\subsection}[hang]{\NormalBaseline\filright\itshape}
{\thesubsection.}
{6pt}
{#1}
[]
\titleformat{\subsubsection}[runin]{\NormalBaseline\filright\itshape}
{\hspace{16pt}\thesubsubsection}
{6pt}
{#1}
[]
\titleformat{\paragraph}[runin]{\NormalBaseline}
{\theparagraph}
{6pt}
{#1}
[]
\titleformat{\subparagraph}[runin]{\NormalBaseline}
{\thesubparagraph}
{6pt}
{#1}
[]

\titlespacing{\section}{0pt}{1.5\baselineskip}{.2\baselineskip}  
\titlespacing{\subsection}{0pt}{1.5\baselineskip}{.2\baselineskip}  
\titlespacing{\subsubsection}{0pt}{1.5\baselineskip}{.2\baselineskip}  
\titlespacing{\paragraph}{0pt}{.5\baselineskip}{10pt}  
\titlespacing{\subparagraph}{0pt}{.5\baselineskip}{10pt}

\usepackage{caption}
\DeclareCaptionLabelFormat{figlabel}{Figure #2}
\date{}
\makeatother

\usepackage{float}

\begin{document}

\title{Bio-Inspired Design of Artificial Striated Muscles Composed of Sarcomere-Like Contraction Units}
\def\RunningHead{
Artificial Myofibril Composed of Multiple APS
}
\def\RunningAuthor{labazanova \etal}
\author{Luiza Labazanova%
\thanks{The Hong Kong Polytechnic University, Department of Mechanical Engineering, Kowloon, Hong Kong.}
, Zeyu Wu\footnotemark[1], ~Zhengping Gu%
\thanks{Delft University of Technology, Faculty of Mechanical, Maritime and Materials Engineering, Delft, The Netherlands.}
, and David Navarro-Alarcon\footnotemark[1]
}
\maketitle


{\begin{onecolabstract}
Biological muscles have always attracted robotics researchers due to their efficient capabilities in compliance, force generation, and mechanical work. 
Many groups are working on the development of artificial muscles, however, state-of-the-art methods still fall short in performance when compared with their biological counterpart. 
Muscles with high force output are mostly rigid, whereas traditional soft actuators take much space and are limited in strength and producing displacement. 
In this work, we aim to find a reasonable trade-off between these features by mimicking the striated structure of skeletal muscles. 
For that, we designed an artificial pneumatic myofibril composed of multiple contraction units that combine stretchable and inextensible materials.
Varying the geometric parameters and the number of units in series provides flexible adjustment of the desired muscle operation. 
We derived a mathematical model that predicts the relationship between the input pneumatic pressure and the generated output force.
A detailed experimental study is conducted to validate the performance of the proposed bio-inspired muscle.

\def\keywordstitle{Keywords}
\end{onecolabstract}}
 
\begin{multicols}{2}

\section{Introduction}
Robotic systems have undergone considerable changes in recent decades.  Progress in advanced manufacturing and sensor technologies has promoted the piecemeal replacement of the conventional rigid robot structures by deformable ones; Soft robots benefit from a high number of degrees of freedom, intrinsic compliant behavior, and safe interaction with humans and surrounding objects. Plenty of research in soft robotics has been inspired by biological systems, as they are mostly comprised of soft, elastic, and flexible tissues \cite{marchese_onal_rus_2014, xie_domel_an_green_gong_wang_knubben_weaver_bertoldi_wen_2020, seok_onal_cho_wood_rus_kim_2013}. Such structures exhibit a wide range of bending curvatures and high performance in complex unstructured environments.

\begin{figure*}[!ht]
\centering
\includegraphics[width=\textwidth]{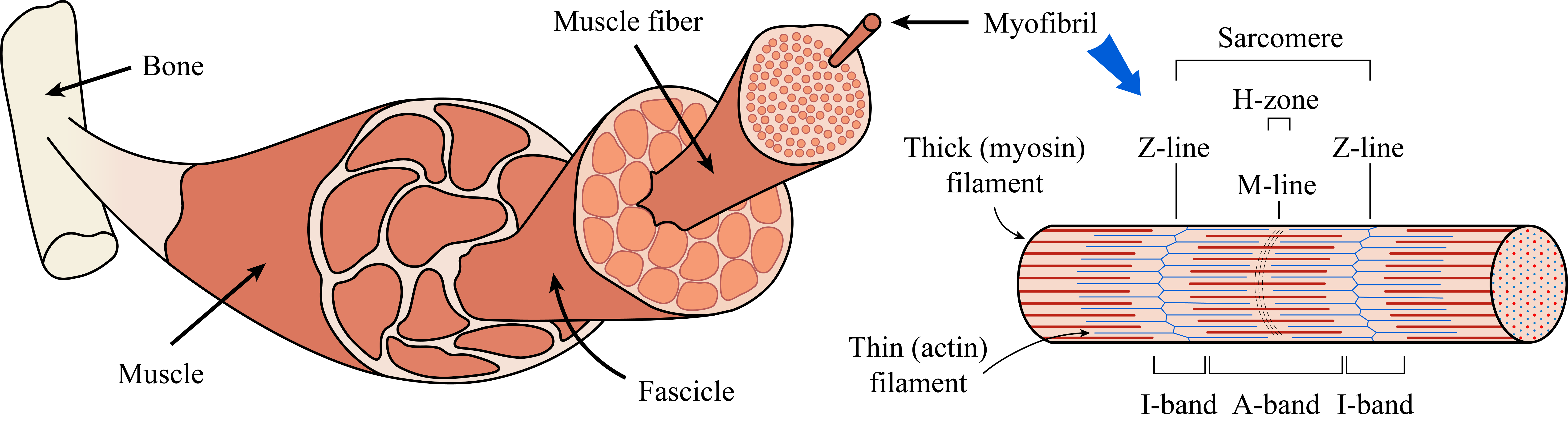}
\caption{Structure of the skeletal muscle.}
\label{fig: muscle-anatomy}
\end{figure*}

Considerable effort has been focused on mimicking the functionality of skeletal muscles through nylon ﬁbers, shape-memory alloys, ionic polymer-metal composites, and others \cite{Mirvakili_Hunter_2017}. Muscles provide large strains, adaptable stiffness, regeneration, and precise grading of force, enabling efficient and optimized motion and load control. Therefore, the development of actuators with performance similar to biological muscles is essential in promoting the ubiquitous application of soft robots in practical tasks. A promising technology is dielectric elastomer actuators (DEAs) that utilize the conversion of electric energy into mechanical work. DEA acts as a compliant capacitor with an insulating elastomer membrane sandwiched between two conductive layers. A voltage applied to electrodes produces electrostatic pressure on elastomer film, inducing large compressive strain. Advantages of DEAs include fast response, substantial actuation strain (> 380\%), and self-sensing capabilities \cite{Pelrine_Kornbluh_Pei_Joseph_2000}. However, DE actuators are driven by very high excitation voltages (> 1 kV) that make them unsafe for human exploitation and vulnerable to failure from the dielectric breakdown \cite{Brochu_Pei_2010}. Less dangerous and well-established artificial muscles are pneumatic artificial muscles (PAMs), often referred to as McKibben muscles. PAMs consist of a flexible tube embedded in a braided mesh clamped at the ends. Upon inflation of the inner chamber by pressurized fluid, radial expansion of the bladder translates into linear contraction. PAMs are easy to fabricate and produce a continuous contraction force, but low compliance limits their performance and nonlinear behavior results in highly complicated analytical modeling \cite{Tondu_2012}. Another type of a pneumatic actuator called HCRPAM was introduced by Han et al \cite{han_kim_shin_2018}. They used a diamond-shaped band to convert the expansion of the elliptical tube in the horizontal direction to the contraction in the vertical direction. HCRPAM showed good results in force and displacement generation, but their drawbacks in comparison with biological muscles include big size and inability to elongate passively under applied loads.

A more compliant alternative of PAMs is fluidic elastomer actuators (FEAs) that have been extensively explored in laboratory conditions \cite{Mosadegh_Polygerinos_2014, Galloway_Polygerinos_Walsh_Wood_2013, Onal_Chen_Whitesides_Rus_2016, Marchese_Katzschmann_2014}. A standard design of FEAs comprises low durometer elastomeric polymers with hollow inner channels. Pressurizing the fluid entrapped in these chambers induces stress in the material, causing its deformation. A motion mode, such as extending, contracting, twisting, or bending, depends on the actuator's geometry and presence of inextensible constraints. Although FEAs have been used in diverse soft manipulators, wearable devices, and adaptive locomotion systems, they are often overlooked in muscle mimicry research due to their expansion nature while the biological muscle is only capable of shortening.

In this work, we introduce a novel design of the artificial muscle inspired by the anatomy of sarcomeres, which are contractile units of striated muscle tissue. The proposed solution consolidates several small FE actuators arranged in a row and connected by soft inextensible links. Elastic properties of FEAs enable passive stretch inherent to biological muscles, whereas pneumatic inflation and the interlinks geometry provide force generation and active shortening of the actuator. 
The main contributions of this article are:
\begin{itemize}
    \item The design and fabrication of a new bio-inspired artificial myofibril composed of multiple small contraction units.
    \item The derivation of an analytical mechanical model for predicting the actuator's behavior.
    \item A detailed experimental study that evaluates the performance of the proposed system.
\end{itemize}

\section{Bio-Inspired Design}
A skeletal muscle is a soft stretchable tissue with a sophisticated striated structure that features repeating functional modules. Muscles are composed of fascicles containing bundles of fibers, themselves composed of parallel bundles of myofibrils, see Fig. \ref{fig: muscle-anatomy}. The latter consists of a series of contractile units, named sarcomeres, composed of arrays of thin (actin) and thick (myosin) filaments, the interaction of which forms the contractile mechanism of the muscle. 
In this work, we aim to imitate this mechanism for the development of a soft actuator.

\begin{figure*}[!ht]
    \centering
    \includegraphics[width=\textwidth ]{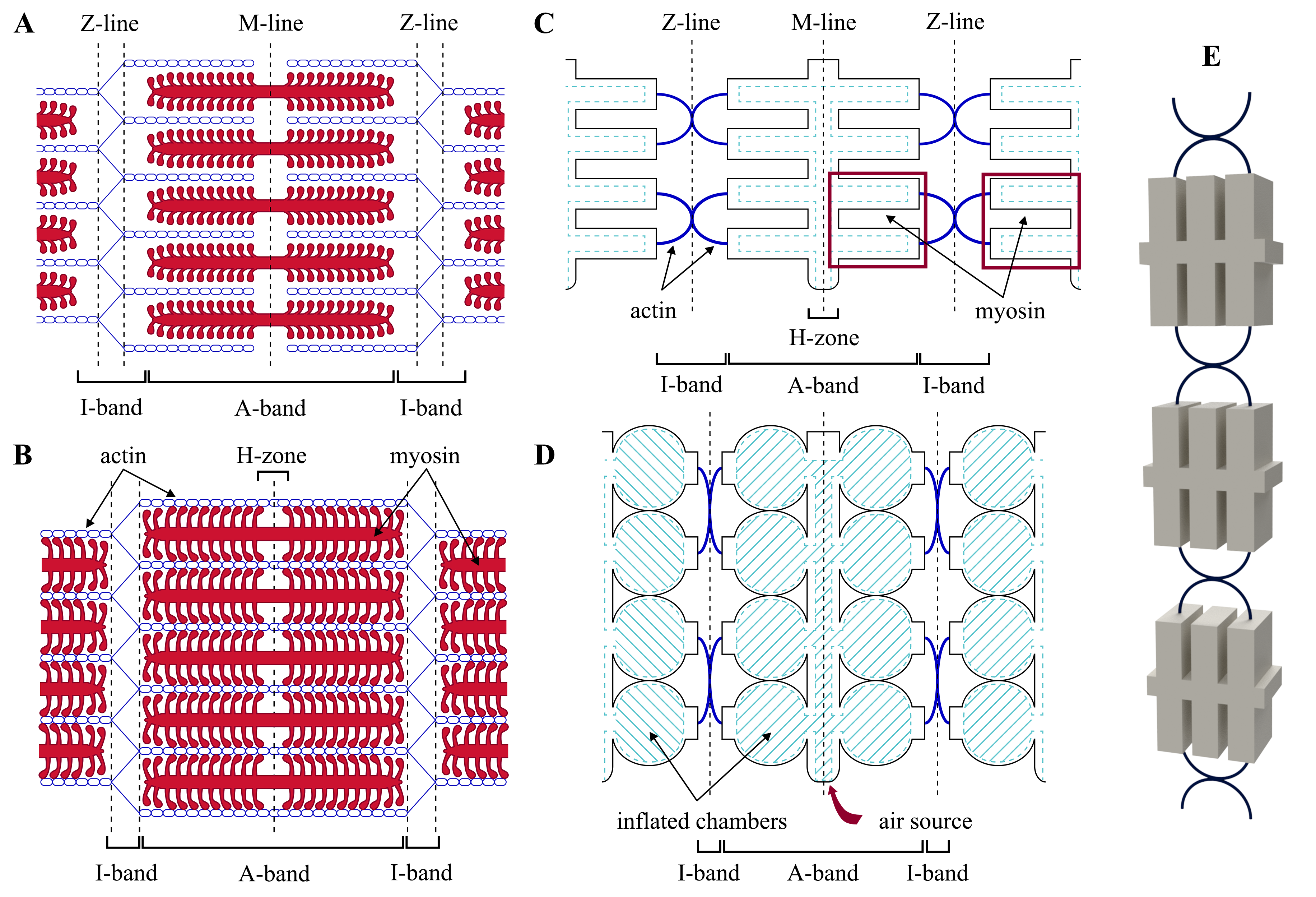}
    \caption{The structure of the biological sarcomere: (A) at the resting state; (B) during contraction. The structure of the artificial pneumatic sarcomere (APS): (C) at the resting state, (D) during contraction. (E) Perspective view of an artificial pneumatic myofibril (APM) composed of three APS.}
    \label{fig: sarcomere-design}
\end{figure*}

The proposed design adopts the bio-inspired structure shown in Fig. \ref{fig: sarcomere-design}A-C. The region of a fiber where only actin filaments are presented is called I-band; The A-band comprises the whole region of the thick filaments. 
There is a gap between the ends of the opposing actin filaments named H-zone. 
In our design, the H-zone corresponds to the inner air channel of the artificial myosin. 
The Z-line anchors the thin filaments to the sarcomere ends; The M-line in the center of the A-band links two myosin filaments together. 
When a muscle (or a motor unit) is activated, myosin heads extend to form cross-bridges with the actin filaments. 
Each cross-bridge acts as a force generator. 
Contraction is produced by actin filaments sliding over the myosin filament, causing narrowing of the I-band and approaching of Z-lines to each other \cite{huxley_hanson_1954, huxley_niedergerke_1954}.

Our design mimics a myofibril (APM) by using a series of artificial pneumatic sarcomeres (APS) formed by soft pneumatic actuators (SPA) connected with flexible ribbons, as shown in Fig. \ref{fig: sarcomere-design}D. 
Each SPA consists of two chambers' arrays pointed in the opposite directions and an air source channel between them. 
For ease of presentation, we will refer to the chambers on one side of the SPA as \emph{myosin} and the inextensible thread attached to them as \emph{actin}. 
Actin ends are inserted in the middle of the outer myosin chambers.  Pressurized air expands fluidic channels embedded within elastomers along the myosin, which increases the distance between ribbon ends.
Consequently, artificial myosins are pulled to each other, reducing the I-band length similar to the sliding filament theory in the biological sarcomere.

\section{Modeling}
\subsection{Geometry of the Artificial Myosin}
\begin{figure*}[!ht]
\floatbox[{\capbeside\thisfloatsetup{capbesideposition={left,top},capbesidewidth=4.2cm}}]{figure}[\FBwidth]
{
    \caption{Geometry of the soft pneumatic actuator (SPA): $t_{w}$ -- thickness of the chamber walls, $a_{ch}$, $b_{ch}$, $h_{ch}$ -- length, width, and height of the inner channels, respectively; $h_{jz}$ -- distance between chambers; $a_{hz}$, $b_{hz}$ -- length and width of the H-zone cross-section, respectively; (A) front view; (B) side view; (C) top view. (D) Geometry of the artificial actin: $l$ -- arc length, $r_{1}$ -- major semi-axis, $r_{1}$ -- minor semi-axis. (E) Schematic representation of the artificial myofibril: $h_{m}$ -- height of a single myosin, $h_{s}$ -- height of a sarcomere, $A$ -- length of the A-band, $I$ -- length of the I-band, $L_{s}$ -- length of a sarcomere, $L_{mf}$ -- length of a myofibril.}
    \label{fig: geometry}
}
{
    \includegraphics[scale=0.195]{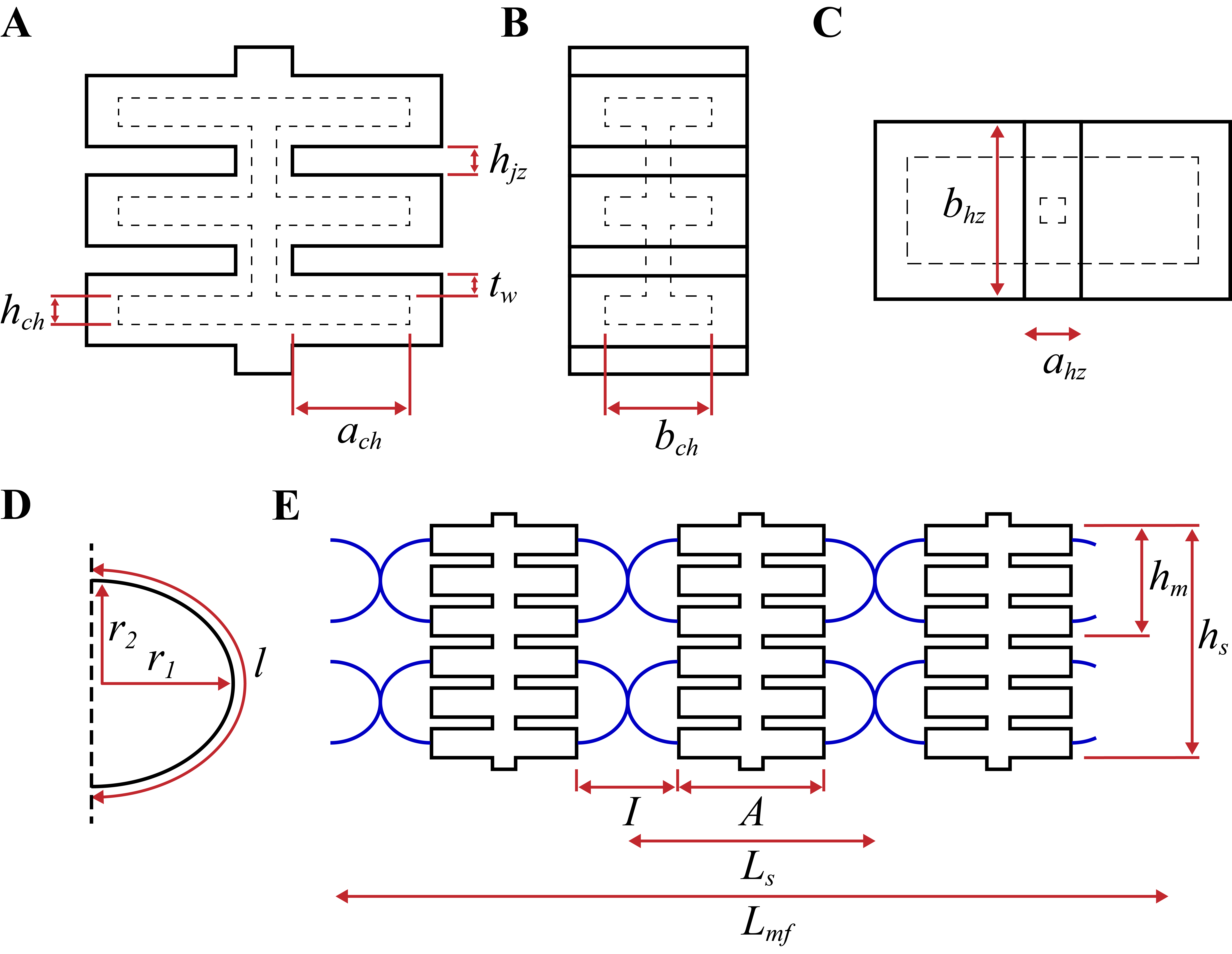}
}
\end{figure*}

Artificial myosin is mimicked by a soft pneumatic actuator with multiple chambers on a side that operate as pneumatic networks (PneuNets) \cite{Mosadegh_Polygerinos_2014}. The number of chambers within the myosin is determined by the desired sarcomere's shortening in a particular case. The H-zone and adjacent myosins are designed as a single unit, see Fig. \ref{fig: geometry}A-C. 
When the air entrapped in the integrated channels is pressurized, the stress generated in the elastomer induces the material to stretch. Inflated walls push each other, increasing the distance between the PneuNets. 

The extent of the myosin deformation depends on fluid pressure and geometrical features of the SPA. Polygerinos et al. explored various parameters of a PneuNets array and defined their impact on the actuator's bending \cite{polygerinos_2013}. In order to analyze the unidirectional elongation of the artificial myosin, a set of geometrical parameters is further identified: The thickness of the chamber walls $t_{w}$; Length $ a_{ch}$, width $b_{ch}$, and height $h_{ch}$ of the inner channels; Distance $h_{jz}$ between PneuNets (junction zone); Length $a_{hz}$ and width $b_{hz}$ of the H-zone's cross-section. 
We use the above features to derive analytical models for myosin elongation and force generation. 

\subsection{Kinematic Analysis}
Figure \ref{fig: geometry}E illustrates the general geometry of an artificial myofibril composed of several sarcomeres. 
The symbol $A$ represents the length of the A-band, and $I$ represents the length of the I-band; $A+I$ defines the total sarcomere's length. $h_{m}$ is the height of a single myosin, and $h_{s}$ is the height of the whole sarcomere. 
When artificial myosins are inflated, $h_{s}$ increases while $I$ decreases. 
When myofibril is passively stretched,  $h_{s}$ slightly decreases whereas $A$ and $I$ increase. 
The initial configuration is chosen to be without any myosin deformation and is denoted with an apostrophe. 
Thus, the resting length of the APM is modeled as:
\begin{equation}
    L_{mf}' = n(A' + I'),
    \label{eq: myofibril-legth-0}
\end{equation}
where $n$ is the number of sarcomeres within the myofibril.

Biological skeletal muscles can be shortened and stretched by approximately 40\% and 70\%, respectively \cite{macintosh_gardiner_mccomas_2006}. Accordingly, we impose the following constraints on the change of the APM length:
\begin{equation}
    0.6 \leq \frac{L_{mf}}{L_{mf}'} \leq 1.7.
    \label{eq: shortening-stretching}
\end{equation}

When artificial myofibril is fully contracted, the distance between sarcomeres almost approaches zero. Therefore, we can establish the relationship between the A-band and I-band as:
\begin{equation}
    \frac{A'}{A'+I'} = 0.6 \implies I' = \frac{2}{3}A'.
    \label{eq: A-I}
\end{equation}

Figure \ref{fig: geometry}D shows the geometry of a single actin, which is assumed to have a half-elliptic shape with arc length $l$. 
The major semi-axis $r_{1}'$ of the ellipse is half of the I-band. 
In the initial configuration, actin is set to form a semi-circle, i.e. $r_{1}' = r_{2}'$. 
Then, from (\ref{eq: A-I}) we have:
\begin{equation}
    l = \frac{\pi}{3}A'.
    \label{eq: actin-arc-length}
\end{equation}

\begin{figure*}[!ht]
    \centering
    \includegraphics[width=\textwidth]{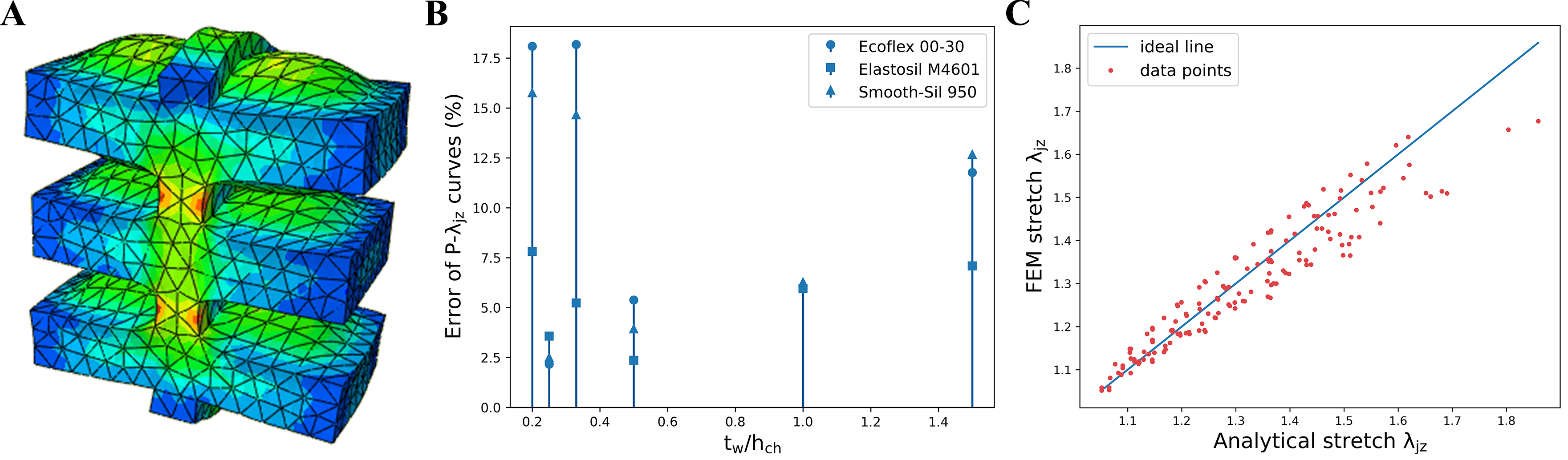}
    \caption{(A) FEM model of inflated myosins. (B) Errors of $P$ -- $\lambda_{jz}$ curves between FEM and analytical models for three different materials. (C) Q-Q plot comparing FEM and analytical stretch $\lambda_{jz}$ of the junction zone.}
    \label{fig: pressure-stretch}
\end{figure*}

When APS is fully contracted, the major semi-axis $r_{1}$ tends to zero and the minor semi-axis $r_{2}$ approaches $\frac{l}{2}$. Since the original height of the myosin is defined as $h_{m} = 2r_{2}' + 2t_{w} + h_{ch}$, its change during inflation can be restricted to:
\begin{equation}
    \frac{2}{3}A' + 2t_{w} + h_{ch} \leq h_{m} \leq \frac{\pi}{3}A' + 2t_{w} + h_{ch}.
    \label{eq: myosin-height}
\end{equation}

The following system can be solved numerically for the accurate determination of the myofibril length at a given deformation $L_{mf}$:
\begin{gather}
    r_{1} \leftarrow \text{solve}
        \begin{cases}
          \text{Semi-ellipse eccentricity}\\
          e = \sqrt{1-\frac{(h_{m}'+\Delta h_{m})^2}{4r_{1}^2}}\\
          \text{Semi-ellipse circumference}\\
          l = r_{1}\int_{0}^{\pi} \sqrt{1-e^2 \sin^2\theta} d\theta \\
        \end{cases} \\
    L_{mf} = n\left(A' + 2r_{1}\right).
    \label{eq: myofibril-length}
\end{gather}

For simplicity of calculation, in further sections we assume that $r_{1}$ and $r_{2}$ are legs of a right triangle with hypotenuse $l/2$.

\subsection{Inflation of the Myosin}
Artificial myosin is an array of PneuNets. According to Marchese and Rus, PneuNets can be modeled as a cylindrical elastomer channel with varying diameter, and with wall thickness depending on the fluid pressure \cite{Marchese_Rus_2015}. 
We extend this model by assuming that the PneuNets have either thin-walled or thick-walled chambers. 
For that, we define the chamber as thin-walled if $t_{w} < \frac{1}{4}h_{ch}$. Thus, stress acting on the walls is determined as $\sigma_{w}=P \cdot K$, where $P$ is the relative fluid pressure and $K$ is either $\frac{h_{ch}}{2t_{w}}$ for the thin-walled chambers or $1+\frac{h_{ch}^2}{2t_{w}(t_{w}+h_{ch})}$ for the thick-walled chambers.

Figure \ref{fig: pressure-stretch} shows that the major myosin elongation occurs between PneuNets. 
The expansion force $F_{e}$ that causes the junction zone to stretch is equal to the force exerted by the walls on each other and can be expressed through the stress $\sigma_{w}$:
\begin{equation}
    F_{e} = 2 \sigma_{w} a_{ch}b_{ch}.
    \label{eq: force-h-zone}
\end{equation}

Since the proposed artificial myosin is built from silicone (a hyperelastic material with non-linear behavior), the stretch of the junction zone can be derived from the strain energy expression. 
The following second-order Yeoh model \cite{yeoh_1993} is used to describe the strain energy function $W$:
\begin{equation}
    W = \sum_{i=1}^{2} C_{i}\left(I_{1}-3\right)^i,
    \label{eq: strain-energy}
\end{equation}
where $C_{i}$ denotes material constants and $I_{1}$ is the first invariant of the three principal stretch ratios $\lambda_i$:
\begin{equation}
    I_{1} = \lambda_{1}^2 + \lambda_{2}^2 + \lambda_{3}^2.
\end{equation}

Polygerinos et al. adapted the above model for PneuNets and derived the following stress formula considering that $\sigma = \frac{\partial W}{\partial \lambda}$, where $\lambda$ is the dominant stretch occurring along the PneuNets \cite{polygerinos_2013}:
\begin{gather}
    W = \sum_{i=1}^{2} C_{i}\left(\lambda^2+\frac{1}{\lambda^2}-2\right)^i, \\
    \resizebox{.8\hsize}{!}{$\sigma = \left(\lambda-\frac{1}{\lambda^3}\right)\left(2C_{1}+4C_{2}\left(\lambda-\frac{1}{\lambda}\right)^2\right)$}.
    \label{eq: cauchy-stress}
\end{gather}

Equation \eqref{eq: cauchy-stress} is employed in the subsequent sections as the procedure $\sigma = \text{cauchyStress}(\lambda)$. 
Since the stress acting on the cross-sectional area of the junction zone can be determined using the force $F_{e}$ from (\ref{eq: force-h-zone}), the relationship between the applied fluid pressure $P$ and stretch $\lambda_{jz}$ occurring in the junction zone can be established as:
\begin{equation}
    \lambda_{jz} \leftarrow \text{cauchyStress}^{-1}\left(\frac{2 \sigma_w a_{ch}b_{ch}}{a_{hz}b_{hz}}\right), 
    \label{eq: cauchy-stress-inverse}
\end{equation}
where the procedure cauchyStress$^{-1}(\sigma)$ is an inverse function of (\ref{eq: cauchy-stress}), which maps a given stress to the corresponding stretch. 

The above model is verified with FEM analysis by using ABAQUS software. 
The dimensions of the SPA model used in the simulation are described as: $a_{ch} = 14$, $b_{ch} = 14$, $a_{hz} = 6$, $b_{hz} = 20$, $h_{jz} = 3$ (all in mm). 
Three hyperelastic materials with different hardness and six $t_{w}/h_{ch}$ ratios are tested. The coefficients $C_i$ applied to the Yeoh model are as follows (all in MPa units): $C_{1} = 0.017$, $C_{2} = -0.0002$, $C_{3} = 0.000023$ for Ecoflex 00-30\cite{steck_qu_kordmahale_2018} with density $\rho = 1070$ kg/m$^3$; $C_{1} = 0.11$,  $C_{2} = 0.02$ for Elastosil M4601\cite{Mosadegh_Polygerinos_2014} with density $\rho = 1130$ kg/m$^3$; $C_{1} = 0.34$ for Smooth-Sil 950\cite{polygerinos_galloway_wang_2019} with density $\rho = 1240$ kg/m$^3$. 
The wall thickness-height ratios of the myosin chambers are set as: $\frac{1}{5}$, $\frac{1}{4}$, $\frac{1}{3}$, $\frac{1}{2}$, $\frac{1}{1}$, and $\frac{3}{2}$. 
Inflating pressure $P$ varied from 0.001 to 0.011 MPa in increments of 0.001 for Ecoflex 00-30; From 0.01 to 0.085 MPa with increments of 0.005 for Elastosil M4601; From 0.02 to 0.22 MPa with increments of 0.02 for Smooth-Sil 950. 

Conducted analysis reveals the linearity of the $P$ -- $\lambda_{jz}$ relationship and high similarity between the actual and estimated data. We applied the normalized Fr\'echet distance \cite{eiter_mannila_1994} to evaluate the discrepancy between analytical and FEM models, see Fig. \ref{fig: pressure-stretch}B. Obtained errors vary from $2.16\%$ to $18.18\%$. The highest errors are observed for smallest $t_{w}/h_{ch}$ ratios, probably, due to the excessive deformation occurring when increasing pressure.
The dispersion of the stretch values retrieved from simulation and obtained by the analytical model is illustrated in Fig. \ref{fig: pressure-stretch}C with R-squared of about 84\%. 

\subsection{Output Force}
\begin{figure*}[!ht]
    \centering
    \includegraphics[width=\textwidth]{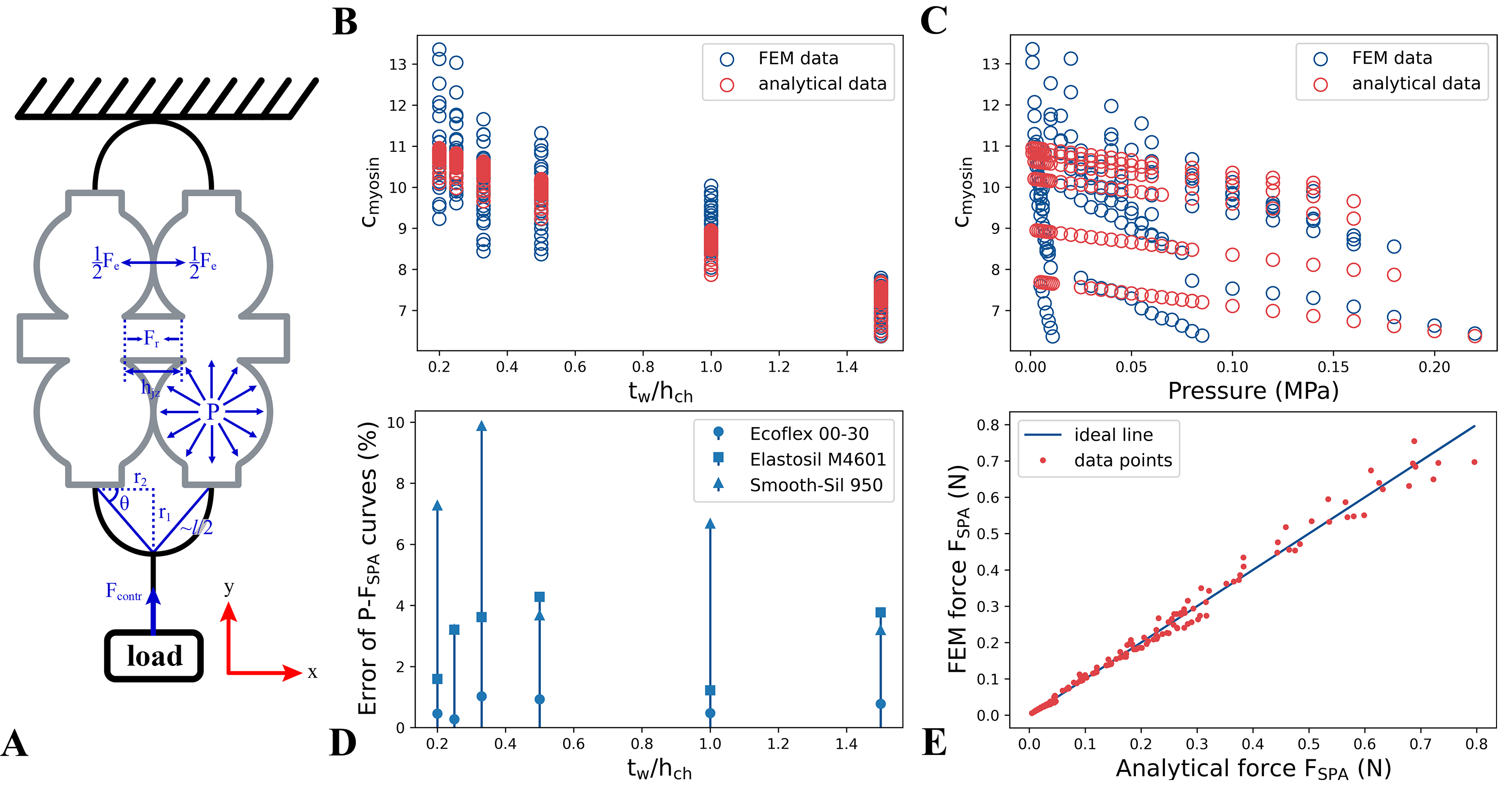}
    \caption{(A) Contraction model of the artificial pneumatic sarcomere. (B) Comparison of the relationships between $t_{w}/h_{ch}$ ratio and FEM/analytical adjustment coefficient $c_{myosin}$. (C) Comparison of the relationships between inflating pressure $P$ and FEM/analytical adjustment coefficient $c_{myosin}$. (D) Errors of $P-F_{SPA}$ curves between FEM and analytical models for three different materials. (E) Q-Q plot comparing FEM and analytical force $F_{SPA}$.}
    \label{fig: pressure-force}
\end{figure*}

The smallest unit of the APS capable of producing force and elongation consists of one H-zone with two pairs of chambers on either side of the junction zone and two actins. 
The number of units depends on the desired muscle shortening.
The contraction model of the sarcomere is designed using FEM simulation. 
This involves the expansion force $F_{e}$ produced by the myosin inflation and contact of the two pairs of lateral walls, and the restoring force $F_{r}$.
The latter force is produced due to the elastic parameters of the material, which tend to restore the myosin to its initial state (Fig. \ref{fig: pressure-force}A). 
The restoring force occurs in the junction zone when it undergoes stretching.

Chou and Hannaford\cite{chou_hannaford_1996} derived a mathematical model for the axial tension of the PAM, based on its cylindrical shape and the principle of virtual work as follows:
\begin{equation}
    F = P\frac{dV}{dL},
    \label{eq: pam-force}
\end{equation}
where $P$ is the relative pressure inside the PAM, $V$ is the PAM volume, and $L$ is its length.

To link the expansion force $F_{e}$ with the SPA deformation, Equation (\ref{eq: force-h-zone}) can be replaced according to \eqref{eq: pam-force}:
\begin{equation}
    F_{e} = 4c_{m}P\frac{dV_{ch}}{dh_{jz}},
    \label{eq: expansion-force}
\end{equation}
where $V_{ch}$ is the volume of a single myosin chamber (assuming that it has a cylindrical form when inflated), $h_{jz}$ is the hight of the junction zone, and $c_{m}$ is an adjustment coefficient. 

Differentiating Equation (\ref{eq: expansion-force}) with respect to stretch of the junction zone $\lambda_{jz}$ gives:
\begin{equation}
    F_{e} = c_{m} P \pi a_{ch} \left( \lambda_{jz} h_{jz} + h_{ch} + 2 t_{w} \right).
    \label{eq: expansion-force-final}
\end{equation}

The restoring force $F_{r}$ can be derived from stress acting along the H-zone in the junction area:
\begin{equation}
    F_{r} = \text{cauchyStress}(\lambda_{jz}) a_{hz} b_{hz}.
\end{equation}

At this step, two forces $F_{e}$ and $F_{r}$ have been determined, which both contribute to the total SPA force denoted as $F_{SPA}$ expressed by (\ref{eq: spa-force}). Since both forces depend on the inflating pressure $P$, the SPA force can be denoted as a function of pressure likewise:
\begin{equation}
    F_{SPA}\left(P\right) = F_{e} - F_{r}.
    \label{eq: spa-force}
\end{equation}

\begin{table*}[!hb]
    \begin{tabular}{ |c|c|c|c| } 
         \hline
         Material & Mean pressure $P$ & Mean max FEM force $F_{SPA}$ & Mean max an. force $F_{SPA}$ \\ 
         \hline
         Ecoflex 00-30 & 0.0055 & 0.0355 & 0.0416 \\ 
         Elastosil M4601 & 0.0401 & 0.2812 & 0.2968 \\ 
         Smooth-Sil 950 & 0.1060 & 0.6957 & 0.7191 \\
         \hline
    \end{tabular}
    \caption{\label{tab: averaged-force}Inflating pressure $P$ and maximum force $F_{SPA}$ averaged over all wall thicknesses.}
\end{table*}

The designed model is validated through comparison of theoretical calculations and FEM analysis. 
Used parameters are the same as for the simulation in the previous section. To determine the adjustment coefficient $c_{m}$ we compare the blocked force obtained from ABAQUS simulation with the analytical model (\ref{eq: spa-force}) and analyze its relation to the myosin parameters such as $t_{w}/h_{ch}$ ratio, material coefficients $C_{1}$ and $C_{2}$ ($C_{3}$ in Ecoflex 00-30 model was omitted), and applied pressure $P$. Graphs in Fig. \ref{fig: pressure-force}B, C show that $c_{m}$ significantly depends on $t_{w}/h_{ch}$ ratio and partially on pressure resulting in correlations $corr(t_{w}/h_{ch},c_{m}) = -0.835642$ and $corr(P,c_{m}) = -0.375539$. Material parameters barely contribute to the change of $c_{m}$: $corr(C_{1},c_{m}) = -0.006081$, $corr(C_{2},c_{m}) = -0.000793$. Therefore, they can be neglected. 

\begin{figure*}[!ht]
    \centering
    \includegraphics[width=\textwidth]{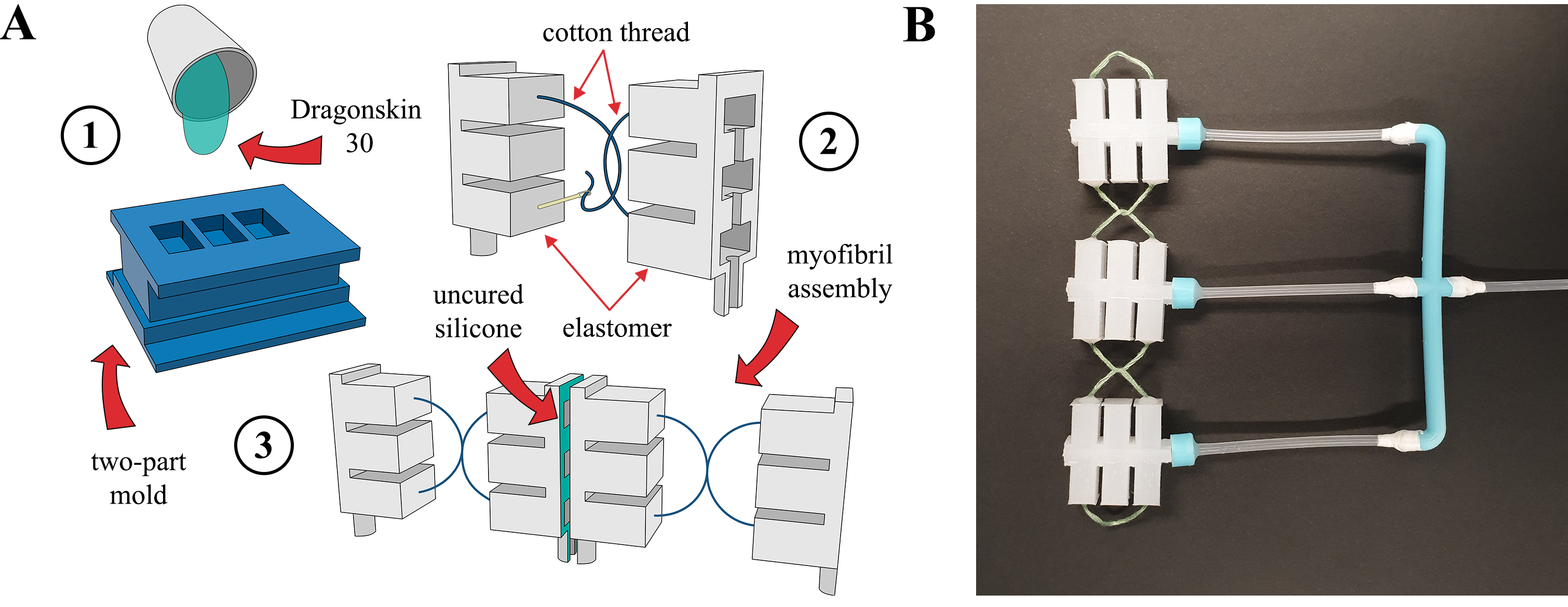}
    \caption{(A) Manufacturing process of an artificial pneumatic myofibril. (B) Assembled myofibril composed of three APS.}
    \label{fig: c-fabrication}
\end{figure*}

By using data fitting, we model the adjustment coefficient $c_{m}$ as follows:
\begin{equation}
    \resizebox{.85\hsize}{!}{$c_{m}\left(\frac{t_{w}}{h_{ch}},P\right) = -2.49\frac{t_{w}}{h_{ch}} - 6.101P + 11.457$}.
    \label{eq: adjustment-coeff}
\end{equation}

The discrepancy between analytical and FEM models evaluated by the normalized Fr\'echet distance between them is shown in Figure \ref{fig: pressure-force}D. Obtained Fr\'echet values are in range of $0.27\% -- 9.88\%$, which is lower than for pressure -- stretch model due to the derived adjustment coefficient. The best match of the curves is observed for Ecoflex 00-30, which is the most stretchable material within tested elastomers. The error increases with decreasing stretchability and does not correlate with the $t_{w}/h_{ch}$ ratio. Figure \ref{fig: pressure-force}E shows the Q–Q (quantile-quantile) plot comparing the FEM and analytical SPA force $F_{SPA}$, which approximately fit each other with R-squared of 98\%. Values of inflating pressure and maximum force produced by actuators from different materials and averaged over all wall thicknesses are listed in Table \ref{tab: averaged-force}. Results reveal that materials with higher strain require less pressure. 
However, materials with higher stiffness are capable of producing larger force; Material selection depends on the required muscle properties. 

Due to constraints produced by actin threads, the SPA force $F_{SPA}$ generated in the $x$ direction is transformed into the sarcomere's contraction force $F_{contr}$ in the $y$ direction (Fig. \ref{fig: pressure-force}A). 
This force can be found from the angle $\theta$ that is formed by $r_{2}$ and the linear approximation of the actin curve $l/2$:
\begin{gather}
    F_{contr} = F_{SPA} \tan{\theta},\\
    \theta = \arccos{\frac{2t_{w} + h_{ch} + 2\lambda_{jz}h_{jz}}{l}}.
    \label{eq: contraction-force}
\end{gather}

\section{Manufacturing Process}
The proposed soft pneumatic actuator that incorporates two myosins and the linking H-zone was built using a molding technique. 
Halves of the SPA were fabricated separately and then glued together. 
A two-part mold with three chambers arranged in a row was designed and 3D-printed from PLA (Inventor, FlashForge). 
Silicone rubber (Dragonskin 30, Smooth-On) was chosen based on the required performance of artificial muscle. 
Dragonskin 30 exhibits a suitable trade-off between stretchability and strength, providing the desired contraction force and sarcomere shortening. 

After elastomer casting and curing, the cotton threads that form artificial actin were inserted into lateral chambers with a needle and crossed with threads of adjacent sarcomeres. 
Finally, holes created by a needle were sealed, and silicone parts were bonded with uncured elastomer. 
Figure \ref{fig: c-fabrication}A illustrates a complete manufacturing process. 
All sarcomeres in a myofibril are connected with flexible tubes to the controllable pneumatic input since they need to contract simultaneously in the same way. 
The assembled artificial myofibril is shown in Fig. \ref{fig: c-fabrication}B.

\section{Results}
\subsection{Experimental Setup}
\begin{figure*}[!ht]
    \centering
    \includegraphics[width=\textwidth]{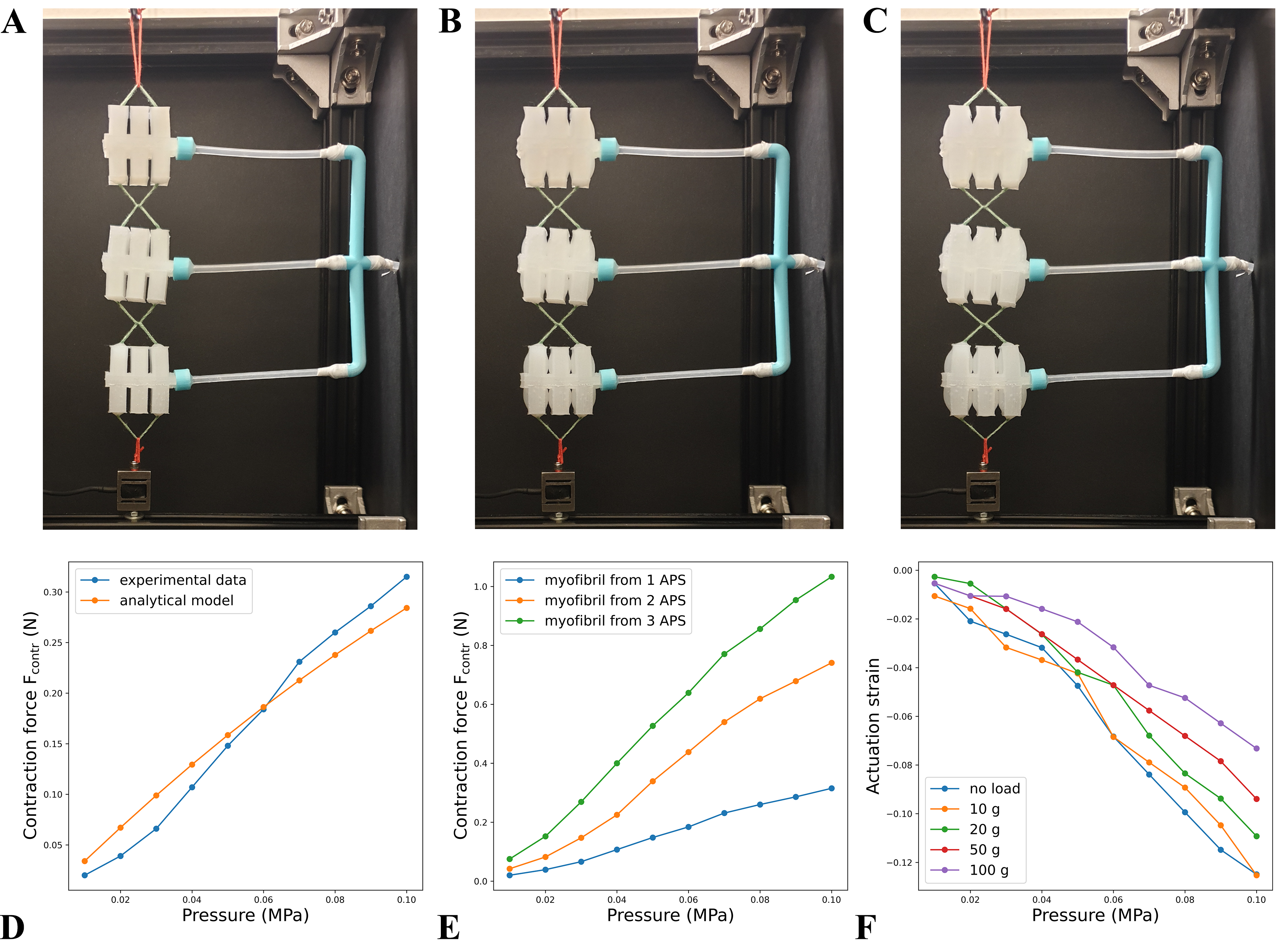}
    \caption{Isometric contraction experiment: (A) resting state; (B) 0.05 MPa; (C) 0.08 MPa. (D) Comparison of $P-F_{contr}$ curves obtained from the analytical model and an isometric contraction experiment for the artificial myofibril composed of one APS. (E) Comparison of $P-F_{contr}$ curves obtained from an isometric contraction experiment on the artificial myofibrils composed of one, two and three APSs. (F) Concentric contraction experiment: applied inflating pressure versus actuation strain $\varepsilon[P,\sigma]$ curves of five different loads on myofibril of three APSs.}
    \label{fig: exp1}
\end{figure*}

\begin{figure*}[!ht]
    \centering
    \includegraphics[width=\textwidth]{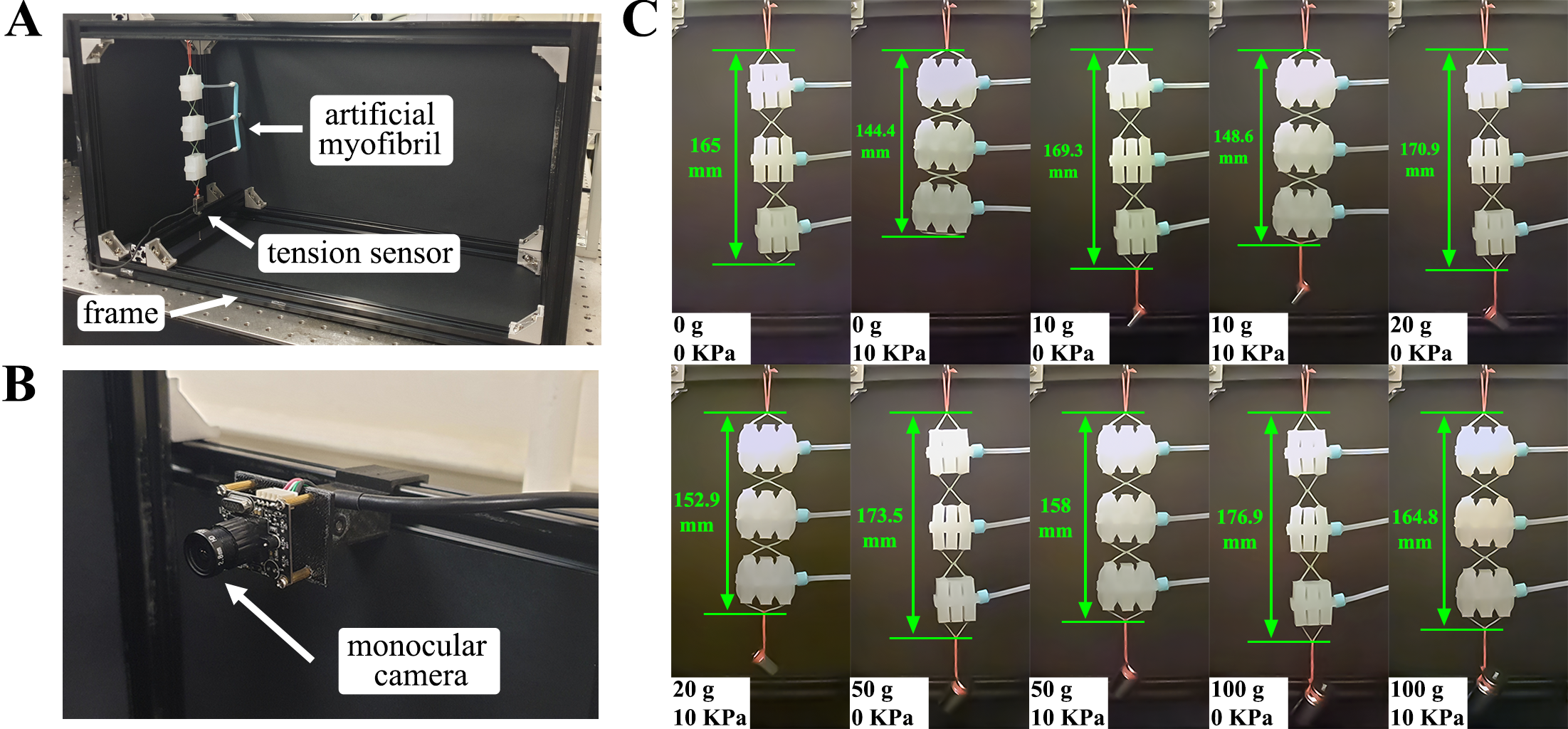}
    \caption{(A) Experimental setup. (B) Camera for detecting the change of myofibril's length. (C) Contraction of the artificial pneumatic myofibril under various loads: 0, 10 g, 20 g, 50 g, and 100 g.}
    \label{fig: exp2}
\end{figure*}

Artificial pneumatic myofibrils composed of several sarcomeres are subjected to experimentation to evaluate their performance during isometric and concentric contractions, and to compare their response with that from the derived analytical model. 
The experimental setup consists of an aluminium frame, proportional pressure regulator (VPPE-3-1-1/8-6-010-E1, Festo), artificial myofibril, tension sensor (JLBS-MD, Bengbu Sensing System Engineering Co., Ltd), and a monocular camera (WX150HD, S-YUE), see Fig. \ref{fig: exp2}A, B. 
Myofibril is suspended from a hook on top of the frame, its opposite end is attached to either a tension sensor or a hanging load. 

According to the derived kinematic model, each sarcomere has the following dimensions (all in mm) at resting state: $A = 30$, $h_{m} = 28$, $a_{ch} = 9.5$, $b_{ch} = 10$, $a_{hz} = 6$, $b_{hz} = 15$, $h_{jz} = 2$, $t_{w} = 1.5$, $h_{ch} = 5$, $l = 32$, with angle $\theta = 51.32^{\circ}$. 
The coefficients $C_{1} = 0.096$ and $C_{2} = 0.0095$ MPa are used as material parameters of DragonSkin 30 in the analytical model \cite{chen_liang_wu_yin_xiang_qu_2018}. 
The pressure is varied from 0.01 to 0.1 MPa with increments of 0.01 MPa. Higher values are not considered in the experiments in order to prevent APM breakage. 
Sarcomeres are associated to a valve and air source through a plastic connector designed to sustain the original shape of the APM.

\subsection{Isometric Contraction}
An isometric contraction experiment is conducted in order to measure the output force of the APM with respect to the air pressure while its length remained fixed, see Fig. \ref{fig: exp1}A -- C. 
Three myofibrils consisting from one to three sarcomeres are tested. 
The maximum forces exerted by each subjects are equal to 0.315, 0.741, and 1.033 N for myofibrils of one, two and three APSs, respectively.
Reducing the wall thickness \cite{polygerinos_2013} and changing material parameters can increase or decrease these values. 

Figure \ref{fig: exp1}D compares the force $F_{contr}$ obtained via experiment with the force computed with the analytical model for the myofibril composed of one APS. 
The Fr\'echet distance between them is $3.6\%$, which states that, the model expressed by \eqref{eq: expansion-force-final} -- \eqref{eq: contraction-force} can be used to predict the myofibril's output force. However, the discrepancy increases with adding contraction units to the myofibril resulting in $17.25\%$ for two APSs and $18.03\%$ for three APSs. 
The plots in \ref{fig: exp1}E show that the contraction force of the multi-SPA chains grows faster than an equivalent scale of a single APS. For a myofibril of two contraction units, a gain of $13.45\%$ in force generation over doubled output of a single APS is observed. Myofibril of three contraction units produces $19.39\%$ higher force versus correspondent multiple of a single APS force.

The benefit of utilizing a fascicle structure in comparison with the individual actuators has been investigated in many applications including soft robotics \cite{robertson_sadeghi_florez_paik_2017}.  Therefore, the analytical model should be extended relative to the number of the contraction units within a myofibril. The difference between analytical and experimental data may also occur due to a slight discrepancy in geometric values of the real and modeled myofibril arisen from an inaccurate fabrication technique as well as approximations applied in the calculation of the SPA force $F_{SPA}$. Nevertheless, the pressure that can be used for inflation is constrained by the limitation of the maximum myofibril contraction, see Eq. \eqref{eq: shortening-stretching}. Thus, the error is not supposed to exceed the above Fr\'echet distances unless longer chains are employed.

\subsection{Concentric Contraction}
The second experiment is aimed to analyze the lifting capabilities of the artificial myofibril under various loads. A concentric contraction is an essential function of the muscle, and its performance highly characterizes muscle efficiency. Loads of 10 g, 20 g, 50 g, 100 g, and contraction without any load were tested. 
The length of the myofibril of three sarcomeres at resting state is measured in advance and is equal to 165 mm. A monocular camera (Fig. \ref{fig: exp2}B) is used for detecting each step of contraction; The resulting lengths are computed with the OpenCV library. 
Figure \ref{fig: exp2}C shows the myofibril transformation under various loads during no active actuation and when the maximum pressure is supplied. 
From these values, the change of myofibril's length can be described as:
\begin{equation}
    0.875 \leq \frac{L_{mf}}{L_{mf}'} \leq 1.07.
\end{equation}
This result partly matches the one in Eq. \eqref{eq: shortening-stretching}. 
However, we did not examine our muscle unit under extreme conditions in this work. 

When the load is applied, the myofibril undergoes passive stretch due to the hyperelastic properties of the silicone used in fabrication. The changed length is defined as $L_{mf}[0,\sigma]$, where 0 means no actuation, and $\sigma$ is the loading stress induced by a weight applied to the APM. 
The actuation strain $\varepsilon$ is the resulting strain after pressure has been injected; It can be found as:
\begin{equation}
    \varepsilon[P,\sigma] = \frac{L_{mf}[P,\sigma] - L_{mf}[0,\sigma]}{L_{mf}[0,0]}.
\end{equation}

Values of actuation strain for bigger weights are apparently lower than for smaller weights. In addition, according to graphs in Fig. \ref{fig: exp1}F, the pressure-actuation strain curves diverge more significantly with increasing pressure denoting the degradation of the APM's lifting capabilities under heavy loads. 
This behavior is inherent to the biological and artificial actuators. 
The relationship between the inflating pressure and the actuation strain is approximately linear. 
Therefore, it simplifies the design of an automatic control strategy for this type of APM.

\section{Conclusion}
We proposed a design method of an artificial muscle that aims to mimic the contraction mechanism of its striated structure. Natural skeletal muscles leverage an arrangement of multiple single actuators in a group, which provides sufficient flexibility in its operation and diversity in shortening and force generation. In this work, we developed an artificial myofibril composed of multiple contraction units named artificial pneumatic sarcomeres. These units deploy the expansion behavior of soft pneumatic actuators to decrease the distance between the units and exert force. One of the additional advantages of using an array of contraction units is that the breakage of one or several of them will only decrease the overall muscle performance avoiding its complete collapse. A concentric contraction experiment demonstrated the performance of the muscle during the passive stretch and active actuation.  

We derived an analytical model capable of predicting an output force of the myofibril based on its geometry parameters and inflating pressure. Experimental results were presented in order to validate the model and the general concept of the APM. The analytical model can be further extended for a better estimation of force and shortening. Improvement of fabrication techniques can help to minimize more the size of contraction units, which can facilitate the integration of myofibrils into common motor units as in the natural muscles. This approach has the potential in developing highly versatile and flexible artificial muscles.


\section*{Acknowledgements} 
This research work was supported in part by the Research Grants Council (RGC) of Hong Kong under grant 14203917, in part by the Key-Area Research and Development Program of Guangdong Province 2020 under project 76, in part by the Jiangsu Industrial Technology Research Institute Collaborative Research Program Scheme under grant ZG9V, and in part by The Hong Kong Polytechnic University under grants YBYT and ZZHJ.



\bibliographystyle{ieeetr}

\begin{thebibliography}{10}

\bibitem{marchese_onal_rus_2014}
A.~D. Marchese, C.~D. Onal, and D.~Rus, ``Autonomous soft robotic fish capable
  of escape maneuvers using fluidic elastomer actuators,'' {\em Soft Robotics},
  vol.~1, no.~1, p.~75–87, 2014.

\bibitem{xie_domel_an_green_gong_wang_knubben_weaver_bertoldi_wen_2020}
Z.~Xie, A.~G. Domel, N.~An, C.~Green, Z.~Gong, T.~Wang, E.~M. Knubben, J.~C.
  Weaver, K.~Bertoldi, L.~Wen, and et~al., ``Octopus arm-inspired tapered soft
  actuators with suckers for improved grasping,'' {\em Soft Robotics}, 2020.

\bibitem{seok_onal_cho_wood_rus_kim_2013}
S.~Seok, C.~D. Onal, K.-J. Cho, R.~J. Wood, D.~Rus, and S.~Kim, ``Meshworm: A
  peristaltic soft robot with antagonistic nickel titanium coil actuators,''
  {\em IEEE/ASME Transactions on Mechatronics}, vol.~18, no.~5, p.~1485–1497,
  2013.

\bibitem{Mirvakili_Hunter_2017}
S.~M. Mirvakili and I.~W. Hunter, ``Artificial muscles: Mechanisms,
  applications, and challenges,'' {\em Advanced Materials}, vol.~30,
  p.~1704407, Dec 2017.

\bibitem{Pelrine_Kornbluh_Pei_Joseph_2000}
R.~Pelrine, R.~Kornbluh, Q.~Pei, and J.~Joseph, ``High-speed electrically
  actuated elastomers with strain greater than 100\%,'' {\em Science},
  vol.~287, p.~836–839, Feb 2000.

\bibitem{Brochu_Pei_2010}
P.~Brochu and Q.~Pei, ``Advances in dielectric elastomers for actuators and
  artificial muscles,'' {\em Macromolecular Rapid Communications}, vol.~31,
  p.~10–36, Jan 2010.

\bibitem{Tondu_2012}
B.~Tondu, ``Modelling of the mckibben artificial muscle: A review,'' {\em
  Journal of Intelligent Material Systems and Structures}, vol.~23,
  p.~225–253, Feb 2012.

\bibitem{han_kim_shin_2018}
K.~Han, N.-H. Kim, and D.~Shin, ``A novel soft pneumatic artificial muscle with
  high-contraction ratio,'' {\em Soft Robotics}, vol.~5, p.~554–566, Oct
  2018.

\bibitem{Mosadegh_Polygerinos_2014}
B.~Mosadegh, P.~Polygerinos, C.~Keplinger, S.~Wennstedt, R.~F. Shepherd,
  U.~Gupta, J.~Shim, K.~Bertoldi, C.~J. Walsh, and G.~M. Whitesides,
  ``Pneumatic networks for soft robotics that actuate rapidly,'' {\em Advanced
  Functional Materials}, vol.~24, p.~2163–2170, Jan 2014.

\bibitem{Galloway_Polygerinos_Walsh_Wood_2013}
K.~C. Galloway, P.~Polygerinos, C.~J. Walsh, and R.~J. Wood, ``Mechanically
  programmable bend radius for fiber-reinforced soft actuators,'' {\em 2013
  16th International Conference on Advanced Robotics (ICAR)}, Nov 2013.

\bibitem{Onal_Chen_Whitesides_Rus_2016}
C.~D. Onal, X.~Chen, G.~M. Whitesides, and D.~Rus, ``Soft mobile robots with
  on-board chemical pressure generation,'' {\em Springer Tracts in Advanced
  Robotics}, p.~525–540, Aug 2016.

\bibitem{Marchese_Katzschmann_2014}
A.~D. {Marchese}, R.~K. {Katzschmann}, and D.~{Rus}, ``Whole arm planning for a
  soft and highly compliant 2d robotic manipulator,'' in {\em 2014 IEEE/RSJ
  International Conference on Intelligent Robots and Systems}, pp.~554--560,
  2014.

\bibitem{huxley_hanson_1954}
H.~Huxley and J.~Hanson, ``Changes in the cross-striations of muscle during
  contraction and stretch and their structural interpretation,'' {\em Nature},
  vol.~173, p.~973–976, May 1954.

\bibitem{huxley_niedergerke_1954}
A.~F. Huxley and R.~Niedergerke, ``Structural changes in muscle during
  contraction: Interference microscopy of living muscle fibres,'' {\em Nature},
  vol.~173, p.~971–973, May 1954.

\bibitem{polygerinos_2013}
P.~{Polygerinos}, S.~{Lyne}, Z.~{Wang}, L.~F. {Nicolini}, B.~{Mosadegh}, G.~M.
  {Whitesides}, and C.~J. {Walsh}, ``Towards a soft pneumatic glove for hand
  rehabilitation,'' in {\em 2013 IEEE/RSJ International Conference on
  Intelligent Robots and Systems}, pp.~1512--1517, 2013.

\bibitem{macintosh_gardiner_mccomas_2006}
B.~R. Macintosh, P.~F. Gardiner, and A.~J. Mccomas, {\em Skeletal Muscle : Form
  and Function}.
\newblock Human Kinetics, 2006.

\bibitem{Marchese_Rus_2015}
A.~D. Marchese and D.~Rus, ``Design, kinematics, and control of a soft spatial
  fluidic elastomer manipulator,'' {\em The International Journal of Robotics
  Research}, vol.~35, p.~840–869, Oct 2015.

\bibitem{yeoh_1993}
O.~H. Yeoh, ``Some forms of the strain energy function for rubber,'' {\em
  Rubber Chemistry and Technology}, vol.~66, p.~754–771, Nov 1993.

\bibitem{steck_qu_kordmahale_2018}
D.~Steck, J.~Qu, S.~B. Kordmahale, D.~Tscharnuter, A.~Muliana, and J.~Kameoka,
  ``Mechanical responses of ecoflex silicone rubber: Compressible and
  incompressible behaviors,'' {\em Journal of Applied Polymer Science},
  vol.~136, p.~47025, Aug 2018.

\bibitem{polygerinos_galloway_wang_2019}
P.~Polygerinos, K.~Galloway, Z.~Wang, F.~Connolly, J.~T.~B. Overvelde, and
  H.~Young, ``Fiber reinforced actuators: Finite element modelling,'' 2019.

\bibitem{eiter_mannila_1994}
T.~Eiter and H.~Mannila, {\em Computing Discrete Fr\'echet Distance}.
\newblock May 1994.

\bibitem{chou_hannaford_1996}
C.-P. Chou and B.~Hannaford, ``Measurement and modeling of mckibben pneumatic
  artificial muscles,'' {\em IEEE Transactions on Robotics and Automation},
  vol.~12, no.~1, p.~90–102, 1996.

\bibitem{chen_liang_wu_yin_xiang_qu_2018}
Z.~Chen, X.~Liang, T.~Wu, T.~Yin, Y.~Xiang, and S.~Qu, ``Pneumatically actuated
  soft robotic arm for adaptable grasping,'' {\em Acta Mechanica Solida
  Sinica}, vol.~31, p.~608–622, Aug 2018.

\bibitem{robertson_sadeghi_florez_paik_2017}
M.~A. Robertson, H.~Sadeghi, J.~M. Florez, and J.~Paik, ``Soft pneumatic
  actuator fascicles for high force and reliability,'' {\em Soft Robotics},
  vol.~4, p.~23–32, Mar 2017.

\end{thebibliography}

\end{multicols}
\end{document}